\documentclass[10pt,twocolumn,letterpaper]{article}

\usepackage[pagenumbers]{cvpr} 

\usepackage[dvipsnames]{xcolor}
\usepackage{multirow}
\usepackage{graphicx}

\definecolor{cvprblue}{rgb}{0.21,0.49,0.74}
\definecolor{mylilac}{RGB}{200,162,200}
\usepackage[pagebackref,breaklinks,colorlinks,citecolor=cvprblue]{hyperref}
\usepackage{graphicx} %
\usepackage{cuted}
\usepackage{cinzel}
\usepackage[T1]{fontenc} 
\usepackage{comment}
\usepackage{float}
\usepackage[titletoc]{appendix}

\usepackage{amssymb}
\usepackage{pifont}
\newcommand{\cmark}{\ding{51}}%
\newcommand{\xmark}{\ding{55}}%

\newcommand{\spot}{SPOT\xspace}
\newcommand{\posiv}[1]{\textcolor{teal}{#1}}
\newcommand{\negav}[1]{\textcolor{magenta}{#1}}

\title{\spot! Revisiting Video-Language Models for Event Understanding}

\author{
\textbf{Gengyuan Zhang \textsuperscript{1,2}\thanks{Equal Contribution.} \quad Jinhe Bi \textsuperscript{1}\footnotemark[1] \quad Jindong Gu \textsuperscript{3} \quad Yanyu Chen \textsuperscript{1} \quad Volker Tresp \textsuperscript{1,2}}  \\
\textsuperscript{1} LMU Munich, Munich, Germany\\
\textsuperscript{2} Munich Center for Machine Learning, Munich, Germany \\
\textsuperscript{3} University of Oxford, Oxford, United Kingdom \\
\tt\small zhang@dbs.ifi.lmu.de \quad jinhe.bi@campus.lmu.de \quad jindong.gu@outlook.com \quad tresp@dbs.ifi.lmu.de
}

\begin{document}
\maketitle

\begin{strip}\centering
\vspace{-20mm}
\includegraphics[width=\textwidth, trim=0.2cm 5.1cm 0.2cm 0.5cm, clip]{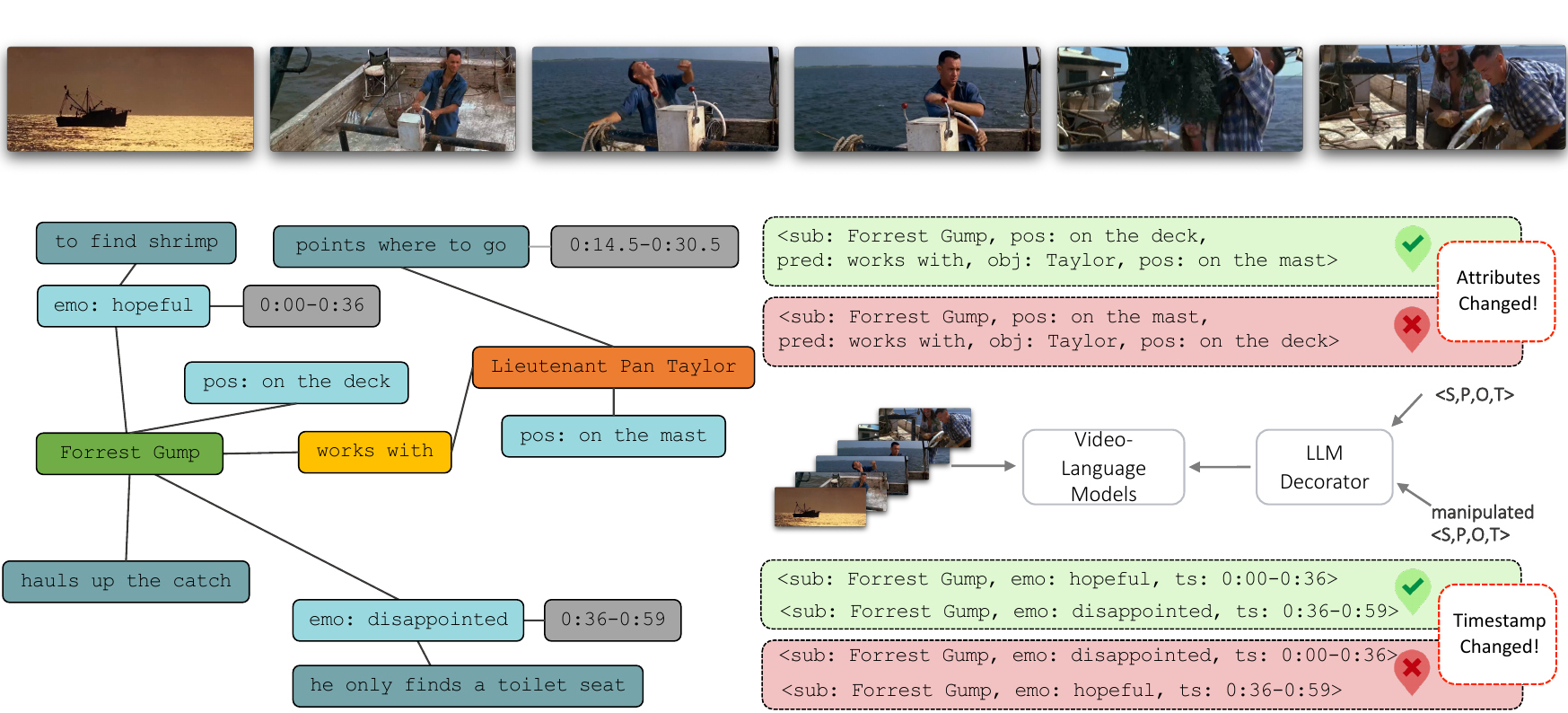}
\mbox{}\\
\captionof{figure}{\textbf{Can Video-Language Models understand events with even only subtle discrepancies?} To explore this, we propose SPOT prober, using extracted event SPOT tuples, <Subject, Predicate, Object, Attribute, Timestamp> from the video scene graph and manipulate event tuples with different manipulation patterns,  like \textit{swapping different entities' attributes} or \textit{swapping predicates of different timestamps}. Based on positive and negative captions generated from these tuples, we aim to reevaluate video-language models' performance sensitivity to any manipulation on event tuples as an indicator of model event understanding abilities.}
\vspace{-0.2cm}
\label{fig:teaser}
\end{strip}

\footnotetext[1]{Preprint version.}

\begin{abstract}

Understanding videos is an important research topic for multimodal learning.
Leveraging large-scale datasets of web-crawled video-text pairs as weak supervision has become a pre-training paradigm for learning joint representations and showcased remarkable potential in video understanding tasks.
However, videos can be multi-event and multi-grained, while these video-text pairs usually contain only broad-level video captions. This raises a question: with such weak supervision, can video representation in video-language models gain the ability to distinguish even factual discrepancies in textual description and understand fine-grained events?
To address this, we introduce SPOT Prober, to benchmark existing video-language models's capacities of distinguishing event-level discrepancies as an indicator of models' event understanding ability. 
Our approach involves extracting events as tuples (<Subject, Predicate, Object, Attribute, Timestamps>) from videos and generating false event tuples by manipulating tuple components systematically. We reevaluate the existing video-language models with these positive and negative captions and find they fail to distinguish most of the manipulated events.
Based on our findings, we propose to plug in these manipulated event captions as hard negative samples and find them effective in enhancing models for event understanding.

\end{abstract}    
\section{Introduction}

Understanding videos is a critical task in vision-language multi-modal learning.
However, comprehending complex object relationships and dynamic attribute changes in videos is challenging.
For example, a short one-minute video in Fig.~\ref{fig:teaser} depicts the relationships and states of multiple characters in a movie and the significant changes in their emotions over time.
Even subtle discrepancies in the video content, such as interchanging characters' positions or altering the chronological sequence of characters' state of mind, can introduce errors in comprehending events. For instance, descriptions like "Forrest Gump on the deck works with Lieutenant Taylor" should be accurate, while "Lieutenant Taylor on the deck works with Forrest Gump" might appear similar but is actually false.

Recently, video-language models have shown significant progress in video understanding tasks. 
By utilizing web-crawled video-text pairs as weak supervision for video-language learning, extensive human annotations and curation are circumvented. Moreover, Video-language models trained by video-text pairs further show exceptional capacities for video understanding tasks~\cite{wang2022internvideo}, making it a popular paradigm for video representation learning.

Even though weak textual supervision showcases generalizable abilities in video understanding at a broad level, we are still concerned regarding whether the video-language models can comprehend such subtle but critical distinctions, as exemplified earlier.

This raises a question: In which granularity and in what aspects can video representations of video-language models distinguish subtle discrepancies within videos? 


To explore this question, we introduce \textit{\spot\space Prober} to benchmark existing video-language models’s capacities of distinguishing event-level discrepancies for indicating models’ event understanding ability. By leveraging structured descriptions of video events in the form of SPOT tuples, "<\textbf{S}ubject, Subject attribute, \textbf{P}redicate, \textbf{O}bject, Object attribute, \textbf{T}imestamp>.", we attempt to extract factual SPOT tuples from video scene graph and generate ``foiled" tuples that contradict facts by manipulating tuple components.

For instance, in Fig.~\ref{fig:teaser}, we generate two untruthful descriptive tuples of the video by swapping the subject and object's position attribute or altering the same entity's emotional attributes. To seamlessly bridge the gap between tuple-formatted descriptions and natural language inputs, we utilize a Large Language Model (LLM) as a decorator, transforming SPOT tuples into coherent textual captions. 

By generating positive and negative captions from SPOT event tuples, we attempt to reevaluate video-language models on a Video-Text Retrieval benchmark by comparing the model performance on positive and negative captions. In this way, we can determine the sensitivity of video-language models to specific manipulations and reveal if the model can distinguish and understand event discrepancies.


Inspired by ~\cite{radenovic2023filtering, momeni2023verbs}, 
we also explore using false captions derived from manipulated SPOT tuples as hard negative samples as post-training to improve video-language models. Our experimental investigations suggest that augmenting with additional negative captions can enhance video-language model representations, particularly evident in downstream tasks.

To summarize, our contributions include:
\begin{itemize}
    \item we propose \spot Prober to reevaluate if video-language models can distinguish subtle but critical video discrepancy;
    by extracting SPOT tuples from the video scene graph and generating truthful(positive) and manipulated(negative) captions, we aim to probe the model sensitivity against different types of manipulation;
    \item we benchmark 5 existing video-language models with \spot Prober on the common Video-Text Retrieval task;
    \item we give in-depth analysis of video-language models' event understanding abilities to distinguish subtle event discrepancies;
    \item we explore using manipulated captions as hard negative samples for post-training, despite improved performance on downstream tasks that demand understanding more structured knowledge of video, the overheads of data annotation and curation are still a hurdle. 
\end{itemize}

\begin{figure*}
    \centering
  \begin{subfigure}{0.42\linewidth}
    \includegraphics[width=\linewidth]{ 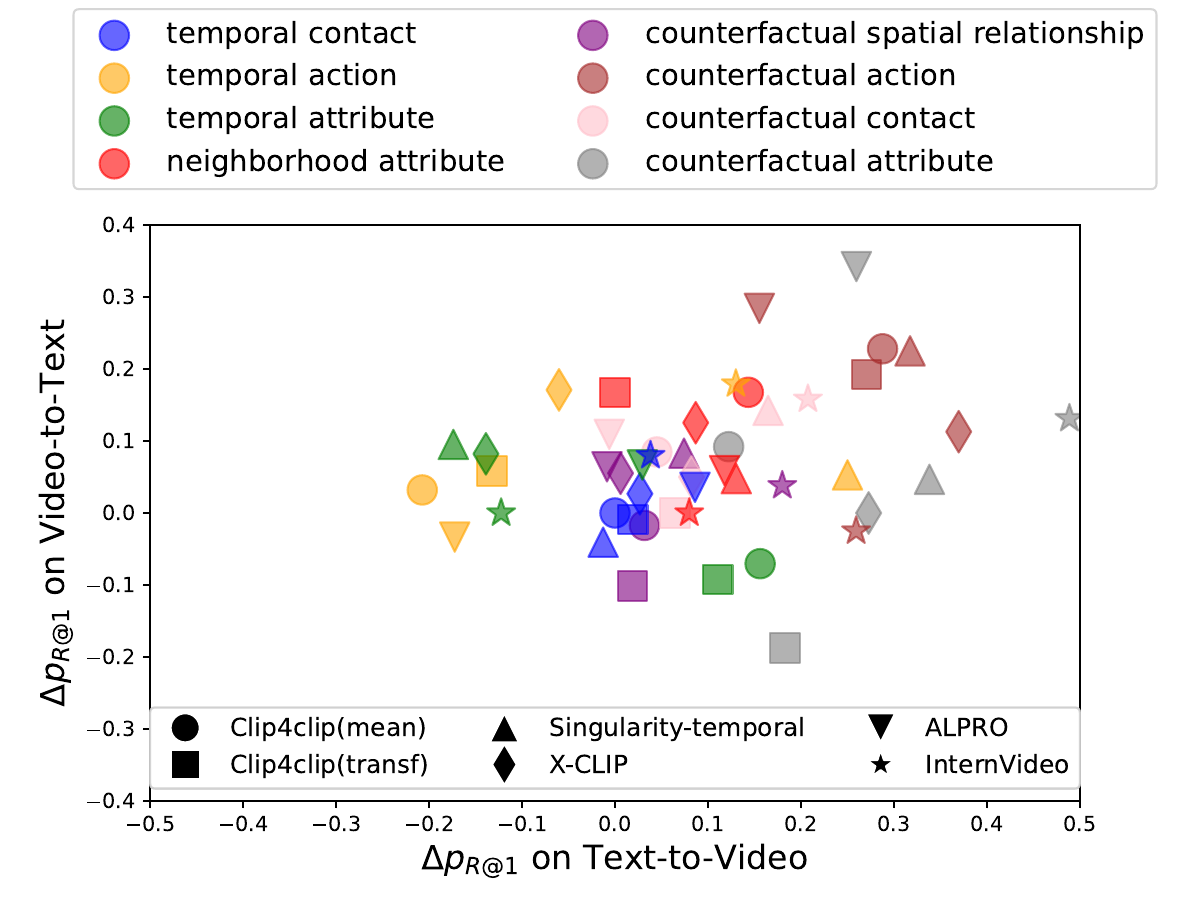}
    \caption{ActivityNet $Recall@1$}
    \label{fig:sub2}
  \end{subfigure}
  \hspace{0.3cm}
    \begin{subfigure}{0.42\linewidth}
    \includegraphics[width=\linewidth]{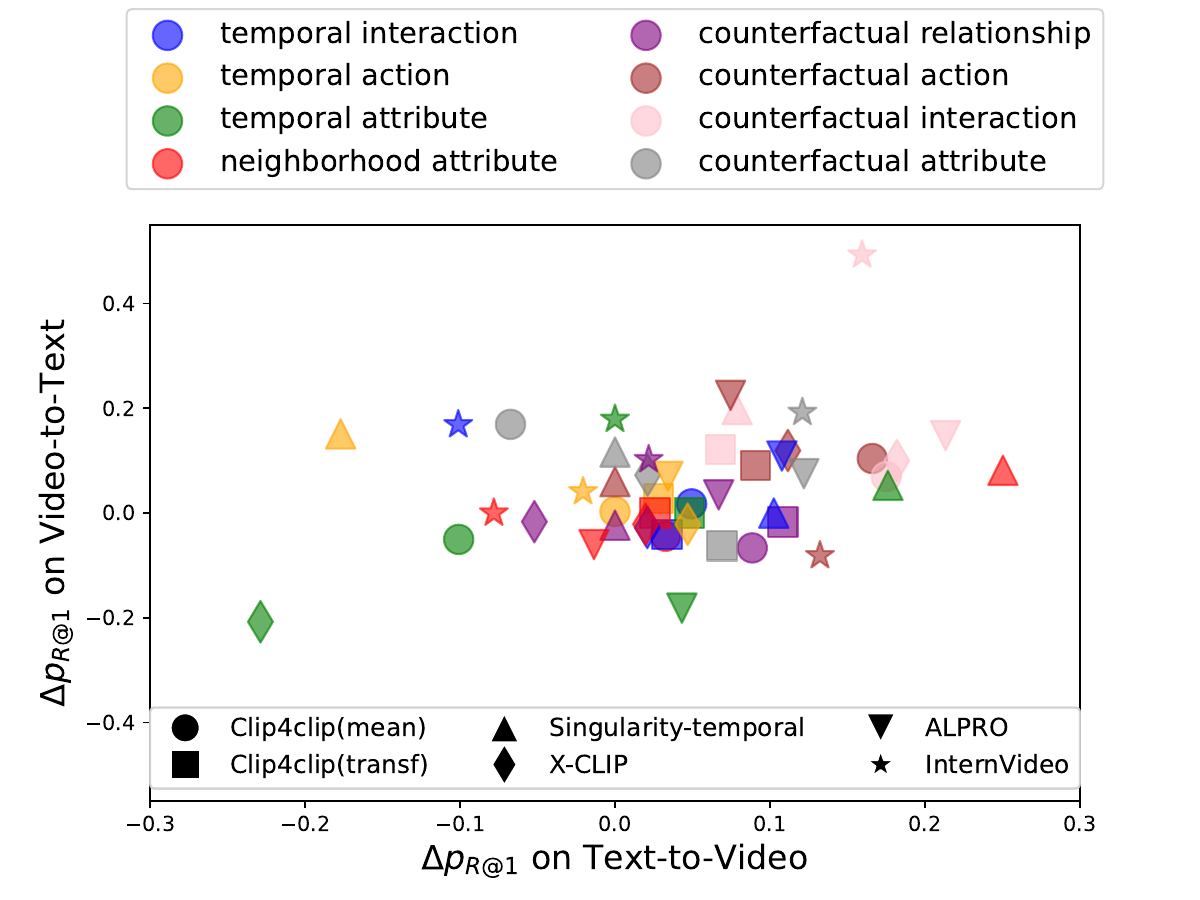}
    \caption{Moviegraph $Recall@1$}
    \label{fig:sub1}
  \end{subfigure}


    \begin{subfigure}{0.42\linewidth}
    \includegraphics[width=\linewidth, trim=0 0 0 2.8cm, clip]{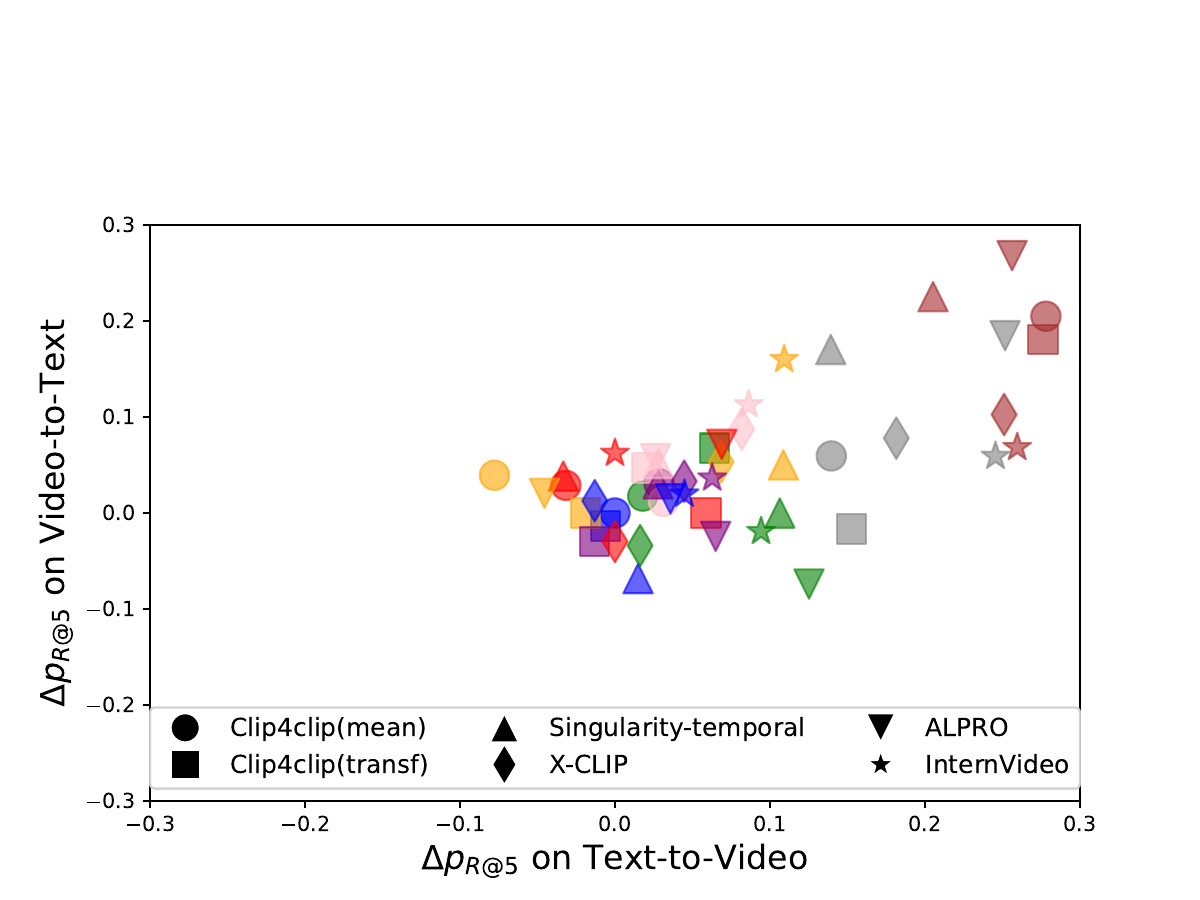}
    \caption{ActivityNet $Recall@5$}
    \label{fig:sub4}
  \end{subfigure}
    \hspace{0.3cm}
   \begin{subfigure}{0.42\linewidth}
    \includegraphics[width=\linewidth, trim=0 0 0 2.8cm, clip]{ 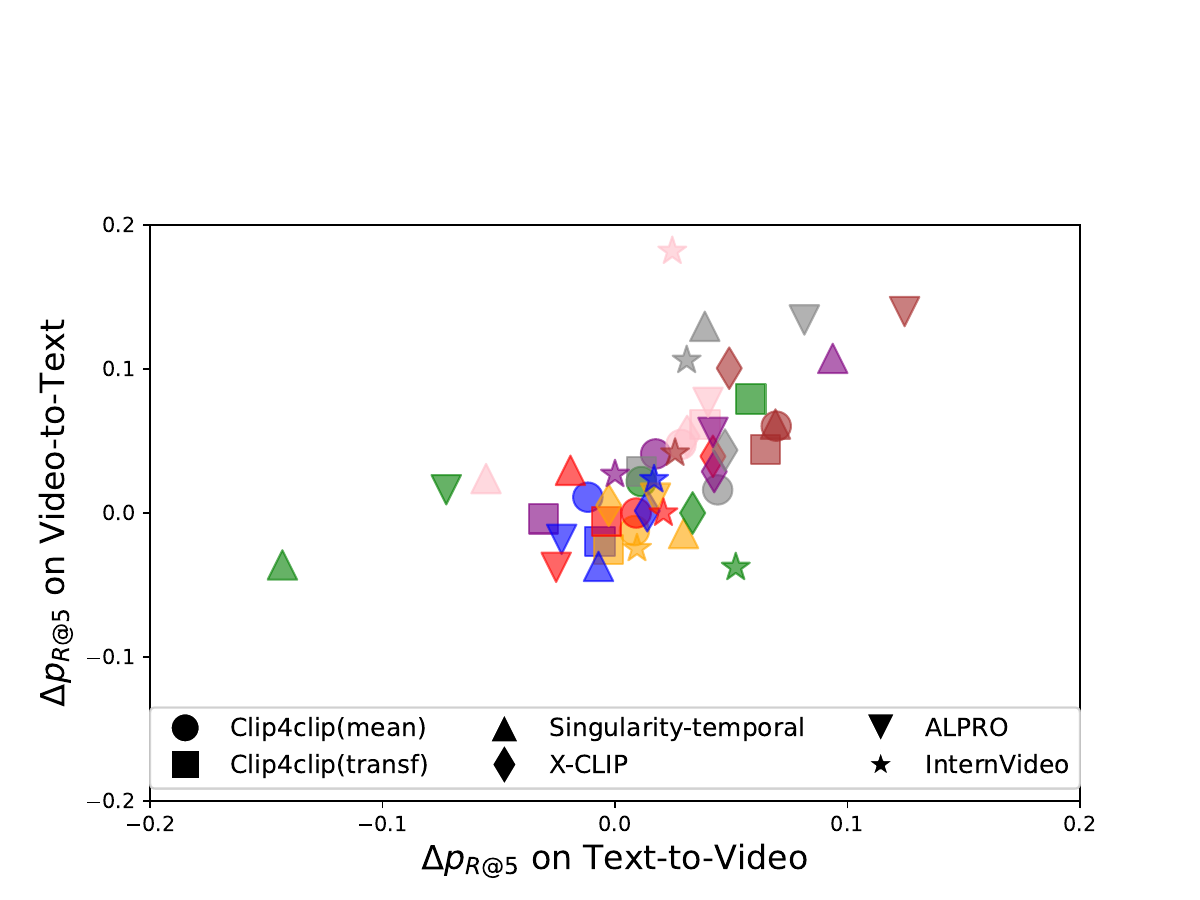}
    \caption{Moviegraph $Recall@5$}
    \label{fig:sub3}
  \end{subfigure}

  \caption{We plot all models' Relative Performance Gap $\Delta p$ of  $Recall@1$ and $Recall@5$ on two datasets. We denote different manipulation types with different colors and models with different shapes. A large positive number indicates the model's sensitivity to the manipulation type and shows the model can more distinguish such manipulation and comprehend the corresponding event.}
  \label{fig:scatter}
\end{figure*}


\section{Related Works}

\vspace{0.2cm}
\noindent
\textbf{Video-Language Model}
By learning a joint cross-modal representation, video-language models bridge the gap between video representation and natural language. 
Works have shown that with weak supervision of video-text pairs~\cite{wang2023internvid, Bain21}, video-language models can learn good joint representations for downstream video understanding tasks like Video-Text Retrieval~\cite{luo2022clip4clip, lei2022revealing, Bain21}, Video Captioning~\cite{sun2019videobert} and Video Question Answering~\cite{wang2023all, li2021align, lei2022revealing}. Recently, video foundation models trained by large-scale video-text data, models like InternVideo~\cite{wang2022internvideo} even manifest excellent generalization performance on various downstream tasks.

\vspace{0.2cm}
\noindent
\textbf{Understanding Vision-Language Model}
To better evaluate model capacities in vision-language representation learning, probing methods have been an important research topic.
SVO probe \cite{hendricks2021probing} designs a benchmark centered on verbs, which assesses subject, verb, and object triplets within the annotated image–sentence pairs in English. This dataset is a valuable resource for evaluating language and visual understanding models.
\cite{yuksekgonul2022and} designs a benchmark to reveal VLM's poor understanding of compositional information, and a composition-aware hard negative mining could improve the performance.
\cite{thrush2022winoground} introduces a novel benchmark to evaluate the ability of vision and language models to conduct visio-linguistic compositional reasoning, and most SOTA models can not outperform chance.
\cite{li2022clipevent} considers image representation event structure information semantic meaning on image-text models.
\cite{ma2023crepe} generates complex negative samples to evaluate VLM's systematicity and productivity and demonstrates that model performance consistently diminishes as novel compositions dominate the retrieval set. \cite{zhao2022vlchecklist} and \cite{cascante-bonilla2023going} evaluate pre-trained VLM's fine-grained understanding abilities of nouns/verbs/attributes. \cite{gandhi2022measuring} introduces a question decomposition engine and designs compositional consistency metrics to reveal that models struggle to reason correctly through most compositions.

For videos, dynamic scenes and multiple events make video representation more complicated and hard to probe. 
\cite{momeni2023verbs} investigates if video-language models can tell the nuance difference between verbs.
\cite{bagad2023test} probes if a video-language model is sensitive to the temporal ordering.
However, these two pioneering works stay on a broad level.

To enhance video-language models' capacities of understanding abilities like verb understanding and compositional understanding,
\cite{chen2020counterfactual} uses generated counterfactual samples in images and texts for the VQA task. \cite{wang2023vilta} uses a text encoder to synthesize complex negative pieces in feature space for ITM loss. These methods shows that by injecting negative samples, vision-language models can improve its representation capacities.


\vspace{0.2cm}
\noindent
\textbf{Video scene graphs in Video-Language dataset}
Video scene graphs represent videos' dynamic and multi-event nature in a graph format.
However, annotating video scene graphs is time and labor-intensive work.
Benefit from the contributions of the community, ~\cite{wu2021star, yu2023anetqa, vicol2018moviegraphs} provide dense and timestamped scene graph-level annotations with rich types of predicates and attributes on top of video data, which enables us to extract SPOT tuples and construct our benchmark.

\begin{table*}[]
\centering
\scriptsize
\setlength{\tabcolsep}{7pt}
\begin{tabular}{llll}
\toprule
Type &
 &
  Positive &
  Negative \\ \hline
temporal contact &
   &
  \begin{tabular}[c]{@{}l@{}}A person with white skin is kneeling on a yellow yard  \\\posiv{before} eventually reaching out to touch the brown mulch \\  \end{tabular} &
  \begin{tabular}[c]{@{}l@{}}A person with white skin is kneeling  on the yellow yard, \\ \negav{while} touching the brown mulch \end{tabular} \\ \hline
temporal action &
   &
   \begin{tabular}[c]{@{}l@{}}
  A silver metal knife is sliding, \\and \posiv{after a while}, a white oven is opened \end{tabular}&
  A silver metal knife is sliding \negav{as} a white oven is being opened \\ \hline
temporal attribute &
   &
  A \posiv{yellow} bike after a while turns \posiv{black} &
  A bike which is \negav{black}, after a while its color is \negav{yellow} \\ \hline
neighborhood attribute &
   &
  The color of the smoke ring is \posiv{white}, and the pipe is \posiv{brown} &
  The smoke ring is \negav{brown} in color, while the pipe is \negav{white} in color \\ \hline
counterfactual contact &
   &
  A white chalk is \posiv{drawn} on the white street &
  The white chalk is \negav{being erased} from the white street \\ \hline
counterfactual action &
   &
  A green cloth punchbag is \posiv{hanging} &
  A green cloth-made punchbag is \negav{swinging} \\ \hline
counterfactual attribute &
   &
  A \posiv{black} bike is parked &
  A \negav{shiny} bike is parked \\ \hline
counterfactual spatial relationship  &
   &
  A boy contestant is \posiv{on a white hill}. &
  A boy contestant is standing \negav{beside the white hill} \\ 
\bottomrule
\end{tabular}
\caption{We showcase the positive and negative captions generated from extracted and manipulated event SPOT tuples on the ActivityNet dataset. We highlight the truthful information in \posiv{cyan} and manipulated information in \negav{magenta}. }
\label{tab:example:ac}
\end{table*}

\section{SPOT Prober: Benchmarking VLMs}\label{sec:method}
\subsection{Preliminary}
Video scene graphs are a structured representation of video content.
Videos can be formulated as a concise form of video scene graphs: $\mathcal{G}:=\mathcal{S}\times\mathcal{A}\times\mathcal{P}\times\mathcal{O}\times\mathcal{A}\times\mathcal{T}$.

Each tuple constitutes an event $\mathcal{E}$ in the video: 

\begin{equation}
    \mathcal{E} = \cup_{t^i} (s, sa, p, o, oa, t^i)
\end{equation}


\subsection{Manipulating SPOT Tuples}
By manipulating the event tuples, we can generate description foils that differ from events in a non-trivial way.
We define \textit{three manipulation methods}: temporal, neighborhood, and counterfactual manipulations.
According to manipulation targets, we can further define \textit{two manipulation patterns}: predicate manipulation and attribute manipulation.

\vspace{0.2cm}
\noindent
\textbf{Temporal Manipulation} is strategically devised to interchange temporal elements in a relative context. This highlights whether the video-language models can sense the temporal ordering of two events, which is a key aspect of videos.

For two events <Subject\textsubscript{1}, 
\text{Subject attribute}\textsubscript{1}, Predicate\textsubscript{1}, Object\textsubscript{1}, Timestamp\textsubscript{1}> and <Subject\textsubscript{2}, 
\text{Subject attribute}\textsubscript{2}, Predicate\textsubscript{2}, Object\textsubscript{2}, Timestamp\textsubscript{2}> in the scene graph, the temporal manipulation $\mathcal{M}_{\text{temporal}}^{\text{predicate}}$ 
and $\mathcal{M}_{\text{temporal}}^{\text{attribute}} $ are defined as

\begin{equation}
\begin{aligned}
 \mathcal{M}_{\text{temporal}}^{\text{predicate}} &= Swap\left((s^1, p^1, o^1, t^1) \cup (s^2, p^2, o^2, t^2)\right)\\&=\left((s^1, p^1, o^1, t^2) \cup (s^2, p^2, o^2, t^1)\right)
\label{temporal-manipulation1}
\end{aligned}    
\end{equation}

\begin{equation}
\begin{aligned}
 \mathcal{M}_{\text{temporal}}^{\text{attribute}} &= Swap\left((s^1, sa^1,t^1) \cup (s^1, sa^2, t^2)\right)\\&=\left((s^1, sa^2, t^1) \cup (s^1, sa^1, t^2)\right)
\label{temporal-manipulation2}
\end{aligned}    
\end{equation}

\vspace{0.2cm}
\noindent
\textbf{Neighborhood Manipulation} is crafted to interchange attributes among distinct neighboring entities in the scene graph. This highlights the ability to connect predicates or attributes with entities.

For one event <Subject\textsubscript{1}, 
\text{Subject attribute}\textsubscript{1}, Predicate\textsubscript{1}, Object\textsubscript{1}, \text{Object attribute}\textsubscript{1}> 
in the scene graph, the neighborhood manipulation $\mathcal{M}_{\text{neighborhood}}^{\text{attribute}}$ is defined as
\begin{equation}
\begin{aligned}
 \mathcal{M}_{\text{neighborhood}}^{\text{attribute}} &= Swap\left((s^1, sa^1,o^1,oa^1)\right) \\
 &= 
    \left((s^1, oa^1,o^1,sa^1)\right)
\end{aligned}
\label{neighborhood-manipulation}
\end{equation}

\vspace{0.2cm}
\noindent
\textbf{Counterfactual Manipulation} is directed towards selecting a counterfactual concept for the particular element within a tuple.

For a single event <Subject\textsubscript{1}, 
\text{Subject attribute}\textsubscript{1}, 
\text{Predicate}\textsubscript{1}, \text{Object}\textsubscript{1}>
in the scene graph, the counterfactual manipulation $\mathcal{M}_{\text{counterfactual}}^{\text{predicate}}$  and $\mathcal{M}_{\text{counterfactual}}^{\text{attribute}}$ are defined as
\begin{equation}
\begin{aligned}
 \mathcal{M}_{\text{counterfactual}}^{\text{predicate}} &= Counter\left(s^1, sa^1, p^1, o^1\right) \\
 &= \left(s^1,sa^1, p^1_{\text{-}}, o^1\right)
\end{aligned}
\label{counterfactual-manipulation1}
\end{equation}

\begin{equation}
\begin{aligned}
 \mathcal{M}_{\text{counterfactual}}^{\text{attribute}} &= Counter\left(s^1, sa^1, p^1, o^1\right) \\
 &= \left( (s^1,sa^1_{\text{-}},  p^1, o^1\right))
\end{aligned}
\label{counterfactual-manipulation2}
\end{equation}
In this context, $sa^1_{\text{-}}$ and $p^1_{\text{-}}$ represent the candidates selected as a counterfactual element.

\begin{table*}[h]
\centering
\scriptsize
\setlength{\tabcolsep}{7pt}
\begin{tabular}{llll}
\toprule
Type &
  &
  Positive &
  Negative \\ \hline
temporal interaction &
   &
  \begin{tabular}[c]{@{}l@{}}Mikael Blomkvist who is male journalist asks Lisbeth Salander, \\ \posiv{before} Lisbeth Salander apologizes Mikael Blomkvist\end{tabular} &
  \begin{tabular}[c]{@{}l@{}}Mikael Blomkvist who is male journalist asks Lisbeth Salander, \\ \negav{after} Lisbeth Salander apologizes Mikael Blomkvist\end{tabular} \\ \hline
temporal action &
   &
  Workers cut branches off a tree, \posiv{before} Workers shred branches &
  Workers cut branches off a tree, \negav{during} Workers shred branches \\ \hline
temporal attribute &
   &
  Tommy is \posiv{excited} after a while Tommy becomes \posiv{skeptical} &
  Tommy is \negav{skeptical} after a while Tommy  \negav{excited} \\ \hline
neighborhood attribute &
   &
   \begin{tabular}[c]{@{}l@{}}
  A male Truck Driver is \posiv{annoyed} \\and the Security Guard looks \posiv{responsible and calm} \end{tabular}&
  \begin{tabular}[c]{@{}l@{}}
  A male Truck Driver looks \negav{responsible and calm} \\while the Security Guard looks \negav{annoyed}.
  \end{tabular} \\ \hline
counterfactual interaction &
   &
  Joey Donner \posiv{talks to} Bianca Stratford, Joey Donner, who is male &
  Joey Donner \negav{flirts with} Bianca Stratford, Joey Donner who is male \\ \hline
counterfactual action &
   &
  Greg Focker \posiv{carries} lawn chairs &
  Greg Focker \negav{assembles} lawn chairs \\ \hline
  
  counterfactual attribute &
   &
  Cecilia is \posiv{honest} woman &
  Cecilia is \negav{irresponsible} woman
 \\ \hline
 counterfactual relationship &
   &
  \begin{tabular}[c]{@{}l@{}}Lisbeth Salander is \posiv{colleague} of Mikael Blomkvist \end{tabular} &
  \begin{tabular}[c]{@{}l@{}}Lisbeth Salander is \negav{the teacher} of Mikael Blomkvist \end{tabular}\\ \bottomrule
\end{tabular}
\caption{We showcase the positive and negative captions generated from extracted and manipulated event SPOT tuples on the MovieGraphs dataset. We highlight the truthful information in \posiv{cyna} and manipulated information in \negav{magenta}.}
\label{tab:example:mg}
\end{table*}

\subsection{Data Preparation}
\vspace{0.2cm}
\noindent
\textbf{Datasets}
We used two video datasets to establish our benchmark: ActivityNet~\cite{caba2015activitynet} and MovieGraphs~\cite{vicol2018moviegraphs}. To extract SPOT tuples, we utilize additional video scene graph annotations in ~\cite{yu2023anetqa, krishna2017dense} to augment the data sources. Datasets statistics can be found in the Tab.~\ref{tab:ds}.
\begin{table}[H]
    \centering
    \footnotesize
    \setlength{\tabcolsep}{0.9pt}
    {\begin{tabular}{cccccc}
    \toprule
         \textbf{Dataset} & \textbf{Domain} & \begin{tabular}[c]{@{}c@{}} \textbf{Video} \\ \textbf{Number} \end{tabular}  & \begin{tabular}[c]{@{}c@{}} \textbf{Average} \\ \textbf{Duration} \end{tabular} & \begin{tabular}[c]{@{}c@{}} \textbf{Total} \\ \textbf{Caption} \end{tabular} & \begin{tabular}[c]{@{}c@{}} \textbf{Average} \\ \textbf{Caption Length} \end{tabular} \\
    \midrule
        ActivityNet\cite{caba2015activitynet} & Activity  & 11.5K& 180s & 16K & 35.11\\
        MovieGraphs\cite{vicol2018moviegraphs} & Movie& 7.6K & 44.28s & 7.6K & 8.1\\
    \bottomrule
    \end{tabular}
    }
    \caption{Statistics of two benchmarks. Original ActivityNet datasets provide the videos in ActivityNet, and the raw captions are provided in ActivityNet-Captions\cite{krishna2017dense}. }
    \label{tab:ds}
\end{table}

\begin{table}[H]
\centering
\footnotesize
\setlength{\tabcolsep}{8pt}
\begin{tabular}{cc}
\toprule
Manipulation Type& Number of Samples \\ \midrule
     temporal contact        &     184    \\
        temporal action     &       62  \\
         temporal attribute    &      102   \\
         neighborhood attribute     &      35   \\
       counterfactual contact      &     1168    \\
      counterfactual action       &      1008   \\
        counterfactual attribute      &      818   \\
        counterfactual spatial relationship     &   935       \\ \bottomrule
\end{tabular}
\caption{Number of extracted SPOT tuples from ActivityNet in each manipulation category in our benchmark.}
\label{tab:num1}
\end{table}

\begin{table}[H]
\centering
\footnotesize
\setlength{\tabcolsep}{8pt}
\begin{tabular}{cc}
\toprule
Manipulation Type& Number of Samples \\ \midrule

      temporal interaction        &    563      \\
        temporal action     &       489   \\
         temporal attribute    &    160     \\
         neighborhood attribute    &    597     \\
         counterfactual relationship    &    583     \\
        counterfactual interaction      &    591     \\
         counterfactual action     &    535      \\
       counterfactual attribute      &   591      \\ \bottomrule
\end{tabular}
\caption{Number of extracted SPOT tuples from MovieGraphs in each manipulation category in our benchmark.}
\label{tab:num2}
\end{table}

\vspace{0.2cm}
\noindent
\textbf{SPOT types}
Depending on the video scene graph annotation difference, we have fine-grained types of predicates and attributes for each dataset.

By applying the above-defined manipulation methods on different tuple components like \textit{Predicate} and \textit{Attribute}, and considering annotation differences in datasets as shown in Tab.~\ref{tab:spottype}, we define 8 different manipulation categories for ActivityNet and MovieGraphs. 

\vspace{0.2cm}
\noindent
\textbf{Manipulation Details}
To control the consistency of the manipulated SPOT tuples and original SPOT tuples, we only manipulate the predicates/attributes of the same fine-grained type as describe in ~\ref{tab:num1}-~\ref{tab:num2}.

\begin{table}[H]
    \centering
\footnotesize
\setlength{\tabcolsep}{8pt}
    {\begin{tabular}{ccc}
    \toprule
         \textbf{Dataset} & \textbf{Predicate Type} & \textbf{Attribute Type}\\
    \midrule
        \multirow{5}{*}{ActivityNet} &  & Occupation  \\ 
        & Interaction & Gender    \\ 
        & Action  &Age \\
        & Relationship &Color \\
          &  &Material \\              
    \midrule
        \multirow{6}{*}{MovieGraphs} & 
        Interaction &Occupation \\
        &  & Gender \\
        & Action &Age \\
        &  &Ethnicity \\  
        & Relationship &Emotion \\  
        &  &Appearance \\  
    \bottomrule
    \end{tabular}}
    \caption{ActivityNet and MovieGraphs define different fine-grained types of predicates and attributes.}
    \label{tab:spottype}
\end{table}

\vspace{0.2cm}
\noindent
\textbf{LLM Decorator}
SPOT tuples are still far from natural descriptions in the video language training corpus.
Given a SPOT tuple, we employ GPT-3.5-turbo as a controlled modifier to convert SPOT tuples into natural languages. 
We use a prompt "\textit{In this task, you are given a sentence, your job is to replace the verb in the sentence with a verb which makes this sentence make sense, please generate 10 sentences.}" and a temperature coefficient to control the sentence quality to avoid introducing misfacts.
Several examples are delineated in Tab.~\ref{tab:example:ac}-~\ref{tab:example:mg}.


\subsection{Benchmark Metrics}
To quantify the video-language models' sensitivity to event discrepancies, we reevaluate the Video-Text retrieval task on the original group of positive captions and the control group of negative captions. 
We believe Video Retrieval tasks are a good measure of learning the similarity of video and text representations based on similarity-based metrics.
We report the \textbf{Relative Performance Gap} $\Delta p$ as follows:
\begin{equation}
    \Delta p = \frac{p - p_{control}}{p}
\end{equation}

where $p$ is the model performance on the original benchmark and $p_{control}$ is the model performance on the control group with manipulated captions.

The motivation behind this is that if the model cannot distinguish the difference between positive and negative descriptions, the performance on the control set with negative samples would drop dramatically, and $\Delta p$ can be considerable. Otherwise, a small or even negative $\Delta p$ indicates that the model cannot distinguish the nuanced difference between positive and negative samples or is distracted by negative captions. 

For all experiments, we report $\Delta p$ on both Video-to-Text and Text-to-Video $Recall@k$, a standard evaluation metric on Video-Text Retrieval. 

\subsection{Baseline Model Choice}
We reevaluate 5 video-language models: Clip4clip~\cite{luo2022clip4clip}, Singularity~\cite{lei2022revealing}, X-CLIP~\cite{ma2022x}, ALPRO~\cite{li2021align}, and InternVideo~\cite{wang2022internvideo}.
The first four are fine-tuned datasets, which we again fine-tune with original videos and captions in ActivityNet and MovieGraphs to ensure the model achieves comparable performance. InternVideo is a large-scale pre-trained video-language model, which we evaluate in a zero-shot setting.

\begin{table*}
\centering
\scriptsize
\setlength{\tabcolsep}{11pt}
\begin{tabular}{lccccc}
\toprule
    \textbf{Model}          & \textbf{Visual Encoder} & \textbf{Textual Encoder} & \textbf{Pre-Training Objectives} & \textbf{Temporal Modeling}  & \textbf{Fine-tuned} \\
\midrule
    Clip4clip(mean)          &      ViT-B/32          &    BERT        &          VTC     &     Mean-Pooling    & \cmark  \\
    Clip4clip(transf)&         ViT-B/32        &   BERT         &       VTC        &        Temporal transformer        & \cmark     \\
    Singularity-temporal        &   BEiT\textsubscript{BASE}            &      BERT      &  VTC+VTM+MLM             &    Temporal transformer (2-layer)      & \cmark    \\
    X-CLIP              &         ViT-B/32        &    BERT      &       VTC        &     Temporal transformer     & \cmark         \\
    ALPRO              &      ViT-B/16          &       BERT\textsubscript{BASE}       &         VTC+VTM+MLM+PEM      &         Temporal transformer+Mean-Pooling      & \cmark      \\
    InternVideo        &      ViT-L/14          &   BERT      &       VTC        &          Linear Projection    & \xmark        \\
\bottomrule
\end{tabular}
\caption{Comparison of model architecture. We compare 5 selected models (6 variants) on different dimensions on: Visual Encoder, Textual Encoder, Pre-training Objectives, Temporal Modeling, and if the model is fine-tuned by us on the datasets.}
\label{tab:modelcomp}
\end{table*}


\begin{table*}[]
\vspace{0.5cm}
\centering
\scriptsize
\setlength{\tabcolsep}{0.7pt}
\begin{tabular}{cccccccccccccccccc}
\toprule
\multirow{3}{*}{Model} &
   & \multicolumn{6}{c}{Temporal}  & 
  \multicolumn{2}{c}{Neighborhood} &
  \multicolumn{8}{c}{Counterfactual}  \\ \cline{3-18} 
  & &
  \multicolumn{2}{c}{contact} &
  \multicolumn{2}{c}{action} &
  \multicolumn{2}{c}{attribute} &
  \multicolumn{2}{c}{attribute} &
  \multicolumn{2}{c}{contact} &
  \multicolumn{2}{c}{action} &
  \multicolumn{2}{c}{attribute} &
  \multicolumn{2}{c}{spatial relationship} \\ \cline{3-18} 
                        &     & $\Delta p_{R@1}$ & $\Delta p_{R@5}$ & $\Delta p_{R@1}$ & $\Delta p_{R@5}$ &  $\Delta p_{R@1}$& $\Delta p_{R@5}$ & $\Delta p_{R@1}$ & $\Delta p_{R@5}$ & $\Delta p_{R@1}$ &  $\Delta p_{R@5}$& $\Delta p_{R@1}$ &$\Delta p_{R@5}$  & $\Delta p_{R@1}$ &$\Delta p_{R@5}$  &$\Delta p_{R@1}$  &$\Delta p_{R@5}$  \\ \hline
\multirow{2}{*}{Clip4clip(mean)}    & T2V & 0.00  & 0.00  & -0.21 &-0.10  &0.16  & 0.02 &0.14  &  -0.03&0.04  & 0.03 &0.29  & 0.28 & 0.12  & 0.14  &0.03  &0.03  \\
                        & V2T & 0.00  & 0.00  & 0.03 & 0.04 &-0.07  &0.02  &0.17  & 0.03 &0.08  &0.01  &0.23  &0.20 &0.09  &0.06  &-0.02  &0.03  \\ \hline
\multirow{2}{*}{Clip4clip(transf)}   & T2V    &0.02    &-0.01    &-0.13  &-0.02  &0.11  &0.64  &0.00  &0.06  & 0.06 & 0.02 &0.27  &0.28  &0.18  &0.15  &0.02  &-0.01  \\
                        & V2T    & -0.01   &-0.01    &0.06  &0.00  &-0.09  &0.07  &0.17  & 0.00 &0.00  &0.05  &0.19  & 0.18 &-0.19  &-0.02  &-0.10  &-0.03  \\ \hline
\multirow{2}{*}{Singularity-temporal}    &  T2V   & -0.01   &0.01    & 0.25 &0.11  &-0.17  &0.11  &0.13  &-0.03  &0.16  &0.03  &0.32  &0.21  &0.34  &0.14  &0.07  &0.03  \\
                        & V2T   &-0.04    & -0.07   &0.05  & 0.05 &0.10  &0.00  & 0.05 &0.04  &0.14  & 0.05 &0.23  &0.23  &0.05  &0.17  &0.08  &0.03  \\ \hline
\multirow{2}{*}{X-CLIP}     & T2V    & 0.03   & -0.01   &-0.06  &0.07  & -0.14 &0.02  &0.09  &0.00  & 0.08 &0.08  & 0.37 &0.25  & 0.27 &0.18  &0.01  &0.04  \\
                        & V2T    &0.03    & 0.01   &0.17  &0.05  &0.08  &-0.03  &0.13  &-0.03  &0.05  &0.09  &0.11  &0.10  &0.00  &0.08  &0.06  &0.03  \\ \hline
\multirow{2}{*}{ALPRO}    & T2V    &0.09    &0.04    &-0.17  & -0.05 &0.03  &0.13  &0.12  &0.07  &-0.01  & 0.03 &0.16  &0.26  &0.26  &0.25  & -0.01 &0.06  \\
                        & V2T    &0.04    &0.01    &-0.03  &0.02  &0.07  &-0.07  &0.06  &0.07  &0.11  &0.06  &0.28  &0.27  &0.34  &0.18  & 0.06 &-0.02  \\ \hline
\multirow{2}{*}{InternVideo} & T2V    & 0.04   & 0.04   &0.13 &0.11  &-0.12  &0.09  &0.08  &0.00  &0.21  &0.09  &0.26  &0.26  &0.49  &0.25  &0.18  &0.60  \\
                        &  V2T   &0.08    & 0.02   &0.18  &0.16  &0.00  &-0.02  &0.00  &0.06  &0.16  & 0.11 & -0.02 &0.07  &0.13  &0.06  &0.04  &0.04  \\
                        \bottomrule
\end{tabular}
\caption{We report the Relative Performance Gap $\Delta p$ on the Video-Text Retrieval task on ActivityNet. A large positive number indicates the model's sensitivity to the manipulation and shows the model can more distinguish such manipulation and comprehend the event.}
\label{tab:result-an}
\end{table*}




\begin{table*}[]
\centering
\scriptsize
\setlength{\tabcolsep}{0.7pt}
\begin{tabular}{cccccccccccccccccc}
\toprule
 
\multirow{3}{*}{Model} &
   & \multicolumn{6}{c}{Temporal} & 
  \multicolumn{2}{c}{Neighborhood} &
  \multicolumn{8}{c}{Counterfactual}  \\ \cline{3-18}
  & &
  \multicolumn{2}{c}{interaction} &
  \multicolumn{2}{c}{action} &
  \multicolumn{2}{c}{attribute} &
  \multicolumn{2}{c}{attribute} &
  \multicolumn{2}{c}{interaction} &
  \multicolumn{2}{c}{action} &
  \multicolumn{2}{c}{attribute} &
  \multicolumn{2}{c}{relationship} \\ \cline{3-18} 
                        &     & $\Delta p_{R@1}$ & $\Delta p_{R@5}$ & $\Delta p_{R@1}$ & $\Delta p_{R@5}$ &  $\Delta p_{R@1}$& $\Delta p_{R@5}$ & $\Delta p_{R@1}$ & $\Delta p_{R@5}$ & $\Delta p_{R@1}$ &  $\Delta p_{R@5}$& $\Delta p_{R@1}$ &$\Delta p_{R@5}$  & $\Delta p_{R@1}$ &$\Delta p_{R@5}$  &$\Delta p_{R@1}$  &$\Delta p_{R@5}$  \\ \hline
\multirow{2}{*}{Clip4clip(mean)}    & T2V & 0.05  & -0.01  &0.00  &0.01  &-0.10  &0.01  &0.03  &0.01  &0.17  &0.03  &0.17  &0.07  &-0.07  &0.04  &0.09  &0.02  \\
                        & V2T & 0.02  & 0.01  & 0.00 & -0.01 & -0.05 &0.02  &-0.04  &0.00  &0.07  &0.05  &0.10  &0.06  &0.17  &0.02  &0.09  &0.02  \\ \hline
\multirow{2}{*}{Clip4clip(transf)}   & T2V    & 0.03   &-0.01  &0.03  &0.00  &0.05  &0.06  &0.03  &0.00  &0.07  &0.04  &0.09  &0.07  &0.07  &0.01  &0.11  &-0.03  \\
                        & V2T    &-0.04    &-0.02    &0.03  &-0.03  &0.00  &0.08  &0.00  &0.00  &0.12  &0.06  &0.09  &0.04  &-0.06  &0.03  &-0.02  &0.00  \\ \hline
\multirow{2}{*}{Singularity-temporal}    &  T2V   & 0.10   & -0.01   &-0.18  &0.03  &0.18  &-0.14  &0.25  &-0.02  &0.08  &-0.06  &0.00  &0.07  &0.00  &0.04  &0.00  &0.09  \\
                        & V2T   &0.00    &-0.04    &0.15  &-0.01  &0.05  &-0.04  &0.08  &0.03  & 0.20 &0.02  &0.06  &0.06  &0.12  &0.13  &0.00  &0.09  \\ \hline
\multirow{2}{*}{X-CLIP}     & T2V    & 0.02   &0.01    &0.05  &0.00  &-0.23  &0.03  &0.02  &0.04  &0.18  &0.03  &0.11  &0.05  &0.02  &0.05  &-0.05  &0.04  \\
                        & V2T    &  -0.03  &0.00    &-0.02  &0.01  &-0.21  &0.00  &-0.02  &0.04  &0.10  &0.05  &0.12  &0.10  &0.07  &0.04  &-0.02  &0.03  \\ \hline
\multirow{2}{*}{ALPRO}    & T2V    & 0.11   &-0.02    &0.03  &0.02  &0.04  &-0.07  &-0.01  &-0.03  &0.21  &0.04  &0.07  &0.12  &0.12  &0.08  &0.07  &0.04  \\
                        & V2T    & 0.11   &-0.02    &0.07  &0.01  &-0.18  &0.02  &-0.06  &-0.04  &0.15  &0.08  &0.22  &0.14  &0.08  &0.13  &0.03  &0.06  \\ \hline
\multirow{2}{*}{InternVideo} & T2V    &-0.10    &0.02    &-0.02  &0.01  &0.00  &0.05  &-0.08  &0.02  &0.16  &0.02  &0.13  &0.03  & 0.12 & 0.03 &0.02  &0.00  \\
                        &  V2T   &0.17    &0.02    &0.04  &-0.02  &0.18  &-0.04  &0.00  &0.00  &0.50  &0.18  &-0.08  &0.04  &0.19  &0.11  &0.10  &0.03  \\
                        \bottomrule
\end{tabular}
\caption{We report the Relative Performance Gap $\Delta p$ on the Video-Text Retrieval task on MovieGraphs. A large positive number indicates the model's sensitivity to the manipulation and shows the model can more distinguish such manipulation and comprehend the event.}
\label{tab:resuls-mg}
\end{table*}

\vspace{0.2cm}
\noindent
\textbf{Clip4clip}\cite{luo2022clip4clip} is built upon the CLIP framework and explores different temporal modeling modules to address video-text retrieval. Our experiments employ two distinct similarity calculators: a parameter-free type, specifically mean pooling, and a sequential type based on the Transformer Encoder. For brevity, we denote them as Clip4clip(mean) and Clip4clip(transf), respectively.

\vspace{0.2cm}
\noindent
\textbf{Singularity}\cite{lei2022revealing} exploits the single-frame bias of video and adopts an extremely sparse sampling method on videos. We use the Singularity-temporal variant described in ~\cite{lei2022revealing} with a 2-layer transformer-based temporal encoder.

\vspace{0.2cm}
\noindent
\textbf{X-CLIP}~\cite{ma2022x} stands out as a multi-grained contrastive model for video-text retrieval. It aims to aggregate fine-grained and cross-grained similarity matrices to achieve instance-level similarity.

\vspace{0.2cm}
\noindent
\textbf{ALPRO}~\cite{li2021align} functions on sparsely-sampled video frames, achieving enhanced cross-modal alignment without relying on explicit object detectors.

\vspace{0.2cm}
\noindent
\textbf{InternVideo}~\cite{wang2022internvideo} serves as a comprehensive video foundation model. It adopts a transformer-based video and text encoder trained on contrastive learning loss and masked vision loss. We use the official implementation and pre-trained model of InternVideo with a CLIP-based Vision Transformer as the video encoder.

A systematic model comparison is listed in Tab.~\ref{tab:modelcomp}.


\subsection{Findings}
We report the relative performance gap $\Delta p$ in Tab.~\ref{tab:result-an}-~\ref{tab:resuls-mg}. To illustrate the numbers in a clearer way, we also plot all models' $\Delta p$ on a scatter plot as shown in Fig.~\ref{fig:scatter}. 

\vspace{0.2cm}
\noindent
\textbf{Performance on Retrieval Benchmark}
In general, we observe that the majority of video-language models exhibit minimal or even negative $\Delta p$ across various manipulation types. The results suggest that video-language models struggle to differentiate or may even be affected by manipulated captions, thereby restricting their proficiency in comprehending events.
Among all the models, ALPRO manifests comparable performance. We speculate that with entity-level alignment, ALPRO can understand fine-grained event better.

\vspace{0.2cm}
\noindent
\textbf{Temporal manipulations are the hardest.}
As demonstrated in Fig.~\ref{fig:scatter}, it is evident that video-language models lack sensitivity to various temporal manipulations (near the origin point). This indicates their limited ability to comprehend events with temporal discrepancies, the chronological order of events, and even the nuanced changes in attributes.

\vspace{0.2cm}
\noindent
\textbf{Conterfactual manipulation is more distinguishable}
As shown in Fig.~\ref{fig:scatter}, models exhibit a significantly higher sensitivity to counterfactual manipulations on predicates and attributes. This observation suggests that video-language models possess a considerable understanding of discrepancies that diverge from the video content.

\vspace{0.2cm}
\noindent
\textbf{Even foundation models cannot distinguish}
We also observe that large-scale pre-trained models like InternVideo~\cite{wang2022internvideo} show low sensitivity to manipulations and cannot distinguish the nuance between positive and negative. It shows that large-scale datasets can benefit learning video representations on a broad level but cannot discriminate more fine-grained events. 

\vspace{0.2cm}
\noindent
\textbf{Understanding Spatial Relationship in Events} is an interesting dimension probing task. We find that on ActivityNet, most models show no sensitivity to spatial relationship manipulation.

\vspace{0.2cm}
\noindent
\textbf{Understanding Abstract}
We notice that dataset annotations involve many abstract concepts in event descriptions like emotion/personality attributes. These attributes undoubtedly are much harder to comprehend and demand video-language models of visual semantic reasoning abilities.
As indicated in Fig.~\ref{fig:scatter}, 

\vspace{0.2cm}
\noindent
\textbf{Out-of-context Event Understanding}
We also notice that scene graph annotations involve complicated out-of-context concepts like name references like "\textit{Forrest Gump}" and relationships like "\textit{the sister of}," which cannot be deduced from the video content directly but demands the video-language model of commonsense knowledge or more comprehensive background information. 

We include a detailed case study in the Supplementary Material for qualitative studies.

\vspace{0.2cm}
\noindent
\textbf{In-video manipulations are harder}
We have demonstrated that existing video-language models cannot distinguish subtle discrepancies, particularly in relation to temporal and neighborhood manipulation within videos. These types of manipulation are undeniably deceptive. Furthermore, we hypothesize that the presence of noisy data of this nature in pre-training datasets may also contribute to this issue.

\section{Learning manipulated events as Hard Negative Samples}
In this section, we explore the feasibility of leveraging negative captions of manipulated \spot tuples as hard negative training, as suggested in \cite{momeni2023verbs,radenovic2023filtering,huang2023structureclip} and assess the model on a fine-grained Video Question Answering task to see if hard negative samples can effectively help video-language models with event understanding.

\subsection{Experiments}
We utilize a two-stage pipeline to post-train the video-language model with hard negative samples. (1) we post-train the video and text encoders using video-text contrastive learning with hard negative samples; (2) we freeze the encoders and incorporate task-specific decoders for downstream tasks.

\vspace{0.2cm}
\noindent
\textbf{Prototype Model} We selected a two-stream structure as our base model, which consists of a ViT-based video encoder and a BERT-based text encoder. This structure is widely utilized in the following works: Clip4clip\cite{luo2022clip4clip} and Singularity\cite{lei2022revealing}. The encoders that are enhanced with post-training using hard negative samples are kept frozen while an additional transformer-based decoder is trained for the QA settings.

\vspace{0.2cm}
\noindent
\textbf{Hard-negative Post-training Loss} 
Video-text Contrastive learning has expeditiously emerged as the predominant methodology for multimodal alignment. 
Drawing inspiration from the work~\cite{momeni2023verbs,radenovic2023filtering,bagad2023test}, we employ the hard negative noise contrastive multimodal alignment loss in this context to perform reweighting of negative samples according to their hardness.
Applied to any given in-batch $X = \{(v^i, t^i)\}_{i=1}^{\mathcal{N}_{\text{in}}}$ with hard negative sample $\{ t^k\}_{k=1}^{\mathcal{N}^i_{\text{gen}}}$
 comprising feature-encoded image-text pairs, our  CL loss enhanced with generated hard negative samples is defined as:

\vspace{-0.4cm}
\begin{equation}
\resizebox{\columnwidth}{!}{
$
\begin{aligned}
\mathcal{L}_{CL}= &-\log \frac{\exp (v^i\cdot{t^i} /\tau)}{
{\exp (v^i\cdot {t^i}/\tau)} + \sum^{\mathcal{N}_{\text{in}}}_{j\neq i}{\exp (v^i\cdot{t^j}/\tau)\omega^{v \rightarrow t}_{i,j}} + \sum^{\mathcal{N}^i_{\text{gen}}}_{k=1}{\exp (v^i\cdot{t^k}/\tau)\omega^{v \rightarrow t}_{i,k}}}
\\
&-\log \frac{\exp (v^i\cdot{t^i} /\tau)}{
{\exp (v^i\cdot {t^i}/\tau)} + \sum^{\mathcal{N}_{\text{in}}}_{j\neq i}{\exp (v^j\cdot{t^i}/\tau)\omega^{t \rightarrow v}_{j,i}}}
\end{aligned}
$
}
\end{equation}

Here $\tau$ represents a temperature parameter, $\mathcal{N}_{\text{in}}$ denotes the set of samples in the batch, and $\mathcal{N}^i_{\text{gen}}$ is the set of generated negative samples based on $t^i$. And weight functions $\omega^{v \rightarrow t}_{i,j}$ and $\omega^{t \rightarrow v}_{j,i}$ are defined as:
\begin{equation}
\resizebox{0.35\textwidth}{!}{
$
\begin{aligned}
&\omega^{v \rightarrow t}_{i,j} = \frac{(\mathcal{N}_{\text{in}} + \mathcal{N}^i_{\text{gen}} - 1) \exp\left(\beta v^i \cdot t^j/\tau\right)}{\sum^{\mathcal{N}_{\text{in}}}_{m\neq i}\exp\left(v^i \cdot t^m/\tau\right)}, \\
&\omega^{t \rightarrow v}_{j,i} = \frac{(\mathcal{N}_{\text{in}} - 1) \exp\left(\beta v^j \cdot t^i/\tau\right)}{\sum^{\mathcal{N}_{\text{in}}}_{m\neq i}\exp\left(v^m \cdot t^i/\tau\right)}
\end{aligned}
$
}
\end{equation}

\vspace{0.2cm}
\noindent
\textbf{Implementation Details}
Inspired by \cite{luo2022clip4clip, lei2022revealing, li2021align}, we use three learning objectives: ITC, ITM, and MLM loss.
We only adopt generated hard negative on ITC loss. We initialize the vision encoder with the BEiT\textsubscript{BASE} \cite{bao2022beit} model, pre-trained on ImageNet-21K \cite{deng20092009}. Meanwhile, the text encoder is initialized with the first 9 layers of BERT\textsubscript{BASE} \cite{devlin2019bert}. Additionally, a 2-layer temporal encoder is incorporated, and we extract four frames for each video. 

\vspace{0.2cm}
\noindent
\textbf{Training Details}
We adopt a two-stage fine-tuning: data-specific post-training and task-specific fine-tuning.
Our optimization strategy spans 10 epochs, utilizing the AdamW \cite{loshchilov2018decoupled} optimizer with an initial learning rate of 1e-4. In the first epoch, a learning rate warm-up is applied, followed by cosine decay \cite{loshchilov2017sgdr} to 1e-6 for the remainder of the training. To expedite the training process, we employ mixed precision training. 
For hard negative post-training, we use a total batch size of 64, and the model is trained on 2 NVIDIA A100 GPUs. For the Question-Answering (QA) task, we keep both encoders from the post-training frozen and only train the text decoder.


\vspace{0.2cm}
\noindent
\subsection{Evaluation}
To better inspect the model's event understanding ability after hard negative training, we evaluate the model on a Compositional Video QA benchmark, AnetQA~\cite{yu2023anetqa}, which features answering fine-grained and temporal questions.

As shown in Tab.~\ref{tab:qa}, injecting manipulated SPOT tuples as hard negative samples is beneficial on Compositional Video QA benchmark. This indicates that leveraging manipulated captions as augmented negative samples can help video-language models with event understanding by learning subtle discrepancies in the video.  

\subsection{Discussion}

\begin{table}
\vspace{0.3cm}
\centering
\footnotesize
\setlength{\tabcolsep}{7pt}
\begin{tabular}{cccc}
\toprule
   & \multicolumn{3}{c}{\textbf{Accuracy} ($\%$)}      \\ \cline{2-4}
             & temporal & counterfactual & overall \\ \midrule
CL loss w/o HN-Samples        &       29.18  &    3.80    &     21.47
\\
CL loss w/ HN-Samples        &       \textbf{33.63}   &    \textbf{5.55}     &     \textbf{22.34}    \\
\bottomrule
\end{tabular}
\caption{The results on the test subset of AnetQA, with and without hard negative, are based on a frozen post-trained Visual and Text encoder, and a trained QA decoder on top. It has been shown that leveraging negative samples from SPOT tuples can improve the performance on fine-grained categories of video questions.}
\label{tab:qa}
\end{table}

\vspace{0.2cm}
\noindent
\textbf{Limitations}
Despite the performance improvement with hard negative sampling, the considerable overhead in scene graph-level video data annotation and curation persists. Given the data scale in pre-training video-language models, the pivotal issue of efficiently providing clean and diverse negative data for video-language pre-training remains an open question.

\vspace{0.2cm}
\noindent
\textbf{Future Direction}
Considering the impractical hard negative mining from video scene graph annotations, and the observation we state that video-language models are more likely to be deceived by temporal and neighborhood manipulation that happens in video, we find it interesting to discover whether we can use temporal masked video prediction as a weak-supervised learning object to guide the video-language models to understand multi-grained and temporal event in videos.

\section{Conclusion}
In this work, we present the SPOT prober, which is designed to assess the event-level discrepancy detection capabilities of existing video-language models, serving as an indicator of their event understanding ability. 

Our methodology involves extracting events from videos as tuples (<Subject, Predicate, Object, Attribute, Timestamps>) and systematically manipulating tuple components to generate false event tuples. By evaluating the existing video-language models using both positive and negative captions generated from these tuples, we discover that they struggle to distinguish the majority of manipulated events like temporally manipulated events; even large-scale pre-trained video-language models cannot survive.
Based on these findings, we propose incorporating these manipulated event captions as hard negative samples, which effectively enhance models for event understanding.

Although it is still impractical to use hard negative samples generated from crowdsourced scene graph annotations, as it requires intensive annotation overhead, we believe this study sheds light on the weaknesses of video-language models in event understanding.


\newpage
{
    \small
    \bibliographystyle{ieeenat_fullname}
    \bibliography{main}
}

\newpage

\appendix

\begin{figure*}[h]
    \centering
    \includegraphics[width=0.98\textwidth, trim=0cm 5.9cm 0cm 4.2cm, clip]{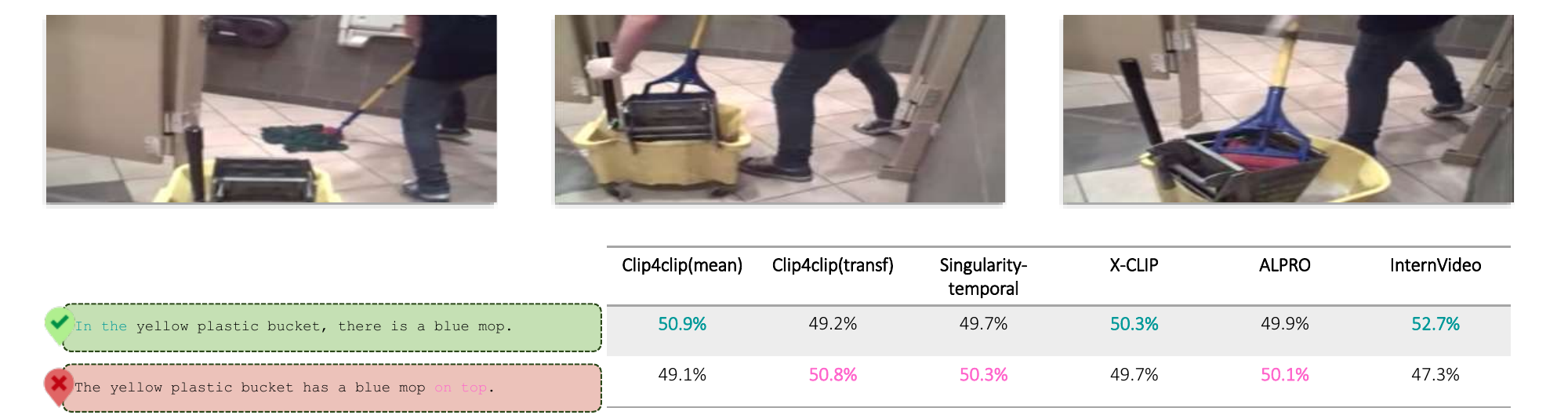}
    \caption{\textbf{Case study: Understanding Spatial Relationship in Events}. This example is sourced from the ActivityNet dataset. In this case, we manipulate the Spatial Relationship to assess the video understanding of each model. We emphasize truthful information in cyan and manipulated information in magenta. Similarly, we annotate the incorrect predictions in \textcolor{magenta}{magenta} and the correct predictions in \textcolor{cyan}{cyan} for clarity. The ``percentage" denotes the relative similarity score calculated from each model.}
    \label{Understanding Spatial Relationship in Events}
    \vspace{-0.4cm}
\end{figure*}

%

In the Supplementary Material, we extend our discussion in Sec.3.6 with a case study in Sec.~\ref{sec:case}.

\section{Case Study}\label{sec:case}
To better reveal the Video-Language Models' event understanding abilities, we conduct a case study of different facets of special cases as we have discussed in Sec.3.6.

\vspace{0.2cm}
\noindent
\textbf{Understanding Spatial Relationship in Events} 
In Figure \ref{Understanding Spatial Relationship in Events}, we present an example to demonstrate the models' capability to understand spatial relationships in events. Specifically, the subtle difference between ``In the ..." and ``... on top" jeopardizes half of the models and shows that they are not sensitive enough to spatial information.

\begin{figure*}
    \centering
    \includegraphics[width=0.98\textwidth, trim=0cm 5.9cm 0cm 4.2cm, clip]{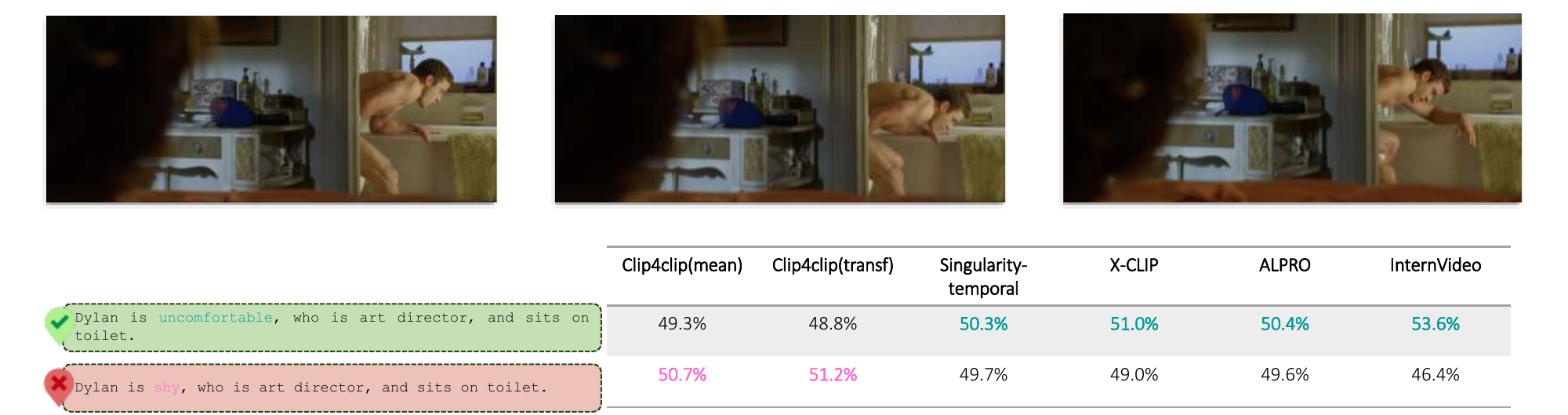}
    \caption{\textbf{Case study: Understanding Abstract}. This example is sourced from the MovieGraph dataset. In this context, we manipulate the emotional attribute of the entity.}
    \label{Understanding Abstract}
    \vspace{-0.4cm}
    
\end{figure*}

\vspace{0.2cm}
\noindent
\textbf{Understanding Abstract} In Fig.\ref{Understanding Abstract}, we present an example to illustrate the models' capability to understand abstract concepts. Specifically, we examine the emotion expressions "\textit{Dylan is uncomfortable}" and "\textit{Dylan is shy}," which present a challenge to the models as they require discerning subtle distinctions between human emotions. These emotions are inherently abstract attributes that necessitate a high level of semantic understanding.

Not all models can understand the subtle discrepancy between them. However, we found that InternVideo as a foundation model, can predict correctly with a higher confidence ($53.6\% vs 46.4\%$).

\begin{figure*}
    \centering
    \includegraphics[width=0.98\textwidth, trim=0cm 5.9cm 0cm 4.2cm, clip]{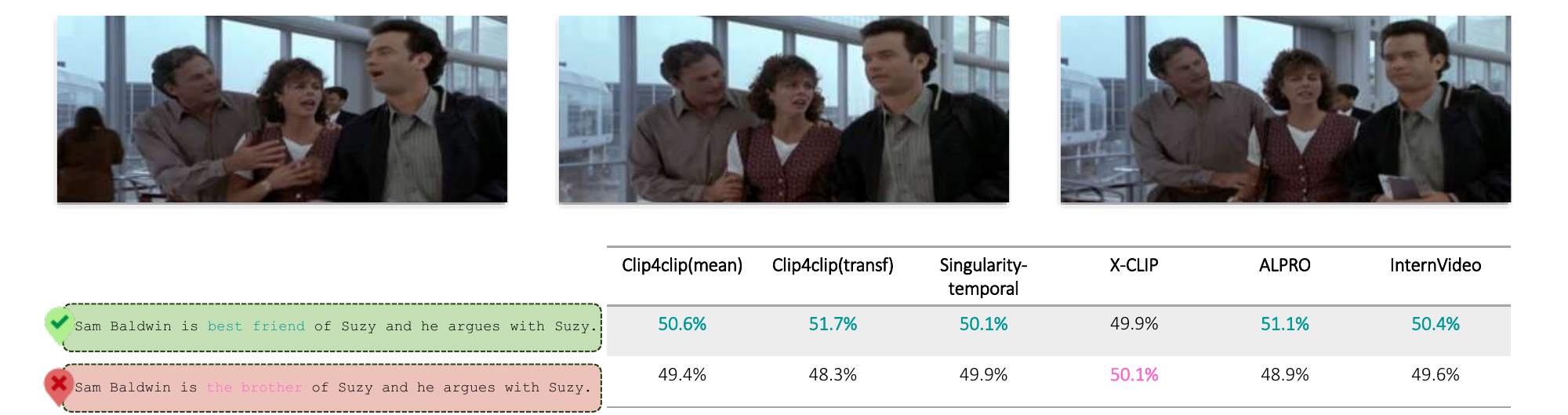}
    \caption{\textbf{Case study: Out-of-context Event Understanding}. This example is also drawn from the MovieGraph dataset. In this instance, we manipulate the relationship between two entities. Even though almost all models make a correct prediction, it cannot be excluded that `\textit{brother}' is also correct.}
    \label{out1}
        \vspace{-0.4cm}
\end{figure*}

\begin{figure*}
    \centering
    \includegraphics[width=0.98\textwidth, trim=0cm 5.9cm 0cm 4.2cm, clip]{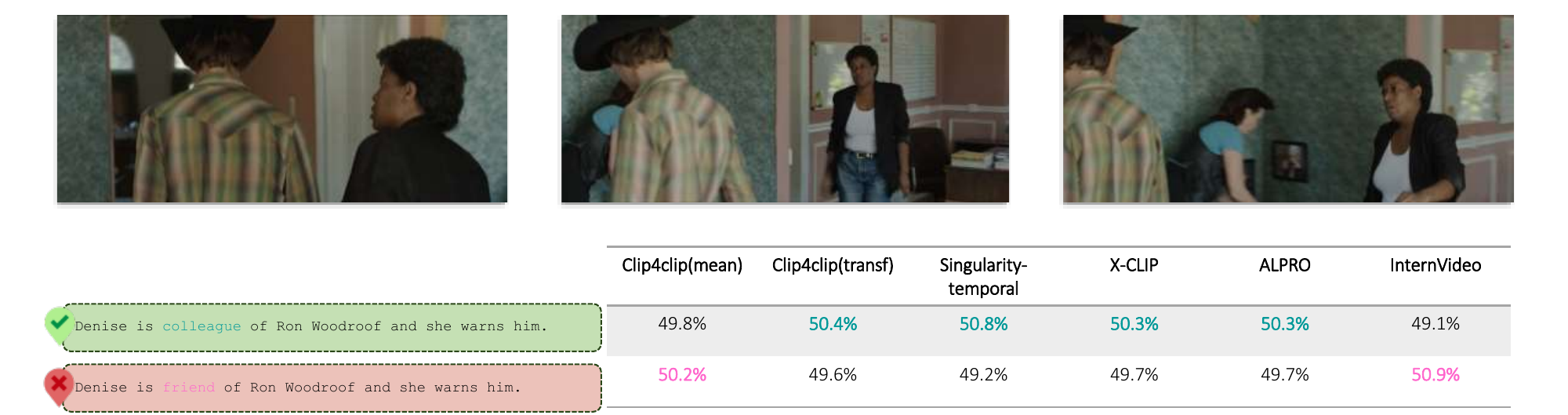}
    \caption{\textbf{Case study: Another example of Out-of-context Event Understanding}s from the MovieGraph dataset. Similarly, `\textit{friend}' and `\textit{brother}' cannot be distinguished without more background information of the characters.}
    \label{out2}
    \vspace{-2cm}
\end{figure*}

\vspace{0.2cm}
\noindent
\textbf{Out-of-context Event Understanding} 
Not all relationships can be deduced by the local visual cues.
In Fig.\ref{out1} and Fig.\ref{out2}, we present two examples to showcase how the models can understand events that are out of context. By "out of context," we mean events that require additional contextual information to be distinguished. For instance, differentiating between a ``\textit{best friend/brother}" relationship and a ``\textit{colleague/friend}" relationship poses a challenge for the model because this distinction cannot be directly inferred from the video data. Even though almost all models make correct predictions, we cannot assert the manipulated captions are totally wrong without additional information.

This actually reveals a blind spot of current video-language models for video understanding: they lack the ability to understand out-of-context knowledge.

\end{document}